\pdfoutput=1

\documentclass[11pt]{article}

\usepackage{emnlp2022}

\usepackage{times}
\usepackage{latexsym}

\usepackage[T1]{fontenc}

\usepackage[utf8]{inputenc}

\usepackage{microtype}

\usepackage{inconsolata}
\usepackage{url}
\usepackage{tikz-dependency}
\usepackage{verbatim}
\usepackage{amsmath}
\usepackage{epsfig}
\usepackage{multirow}
\usepackage{linguex}
\usepackage{graphicx}
\usepackage{subfig}
\usepackage{gb4e}
\usepackage{fullpage}
\usepackage{devanagari}
\usepackage{wrapfig}
\usepackage{cases}
\usepackage{float}
\usepackage{color}
\usepackage{booktabs}
\usepackage{nameref}
\usepackage[titletoc]{appendix}
\usepackage{amsmath}
\usepackage{lipsum}  
\usepackage{color,soul}

\newcommand{\ngram}{\emph{n}-gram}

\title{Discourse Context Predictability Effects in Hindi Word Order}

\author{Sidharth Ranjan \\
  IIT Delhi\\
  \texttt{sidharth.ranjan03@gmail.com} \\\And
  Marten van Schijndel \\
  Cornell University \\
  \texttt{mv443@cornell.edu} \\ \AND
  Sumeet Agarwal \\
  IIT Delhi \\
  \texttt{sumeet@iitd.ac.in} \\\And
  Rajakrishnan Rajkumar \\
  IISER Bhopal \\
  \texttt{rajak@iiserb.ac.in} \\}

\begin{document}
\maketitle
\begin{abstract}
We test the hypothesis that discourse predictability influences Hindi syntactic choice. While prior work has shown that a number of factors (e.g., information status, dependency length, and syntactic surprisal) influence Hindi word order preferences, the role of discourse predictability is underexplored in the literature. Inspired by prior work on syntactic priming, we investigate how the words and syntactic structures in a sentence influence the word order of the following sentences. Specifically, we extract sentences from the Hindi-Urdu Treebank corpus (HUTB), permute the preverbal constituents of those sentences, and build a classifier to predict which sentences actually occurred in the corpus against artificially generated distractors. The classifier uses a number of discourse-based features and cognitive features to make its predictions, including dependency length, surprisal, and information status. We find that information status and LSTM-based discourse predictability influence word order choices, especially for non-canonical object-fronted orders. We conclude by situating our results within the broader syntactic priming literature.
\end{abstract}

\section{Introduction}\label{sect:intro}

Grammars of natural languages have evolved over time to factor in cognitive pressures related to production~\citep{hawkins1994,Hawkins2000} and comprehension~\citep{hawkins04,Hawkins2014}, learnability~\citep{christiansen-chater-2008} and communicative efficiency~\citep{JaegerTily2011,Gibson2019}. In this work, we test the hypothesis that maximization of discourse predictability (quantified using lexical repetition surprisal and adaptive LSTM surprisal) is a significant predictor of Hindi syntactic choice, when controlling for information status, dependency length, and surprisal measures estimated from $n$-gram, LSTM and incremental constituency parsing models. 

Our hypothesis is inspired by a solid body of evidence from studies based on dependency treebanks of typologically diverse languages which show that grammars of languages tend to order words by minimizing dependency length~\citep{Liu2008,futrell2015} and maximizing their trigram predictability~\citep{GildeaJaeger2015}. Parallel to this line of work on sentence-level word order, another strand of work has focused on discourse-level estimates of entropy starting from the Constant Entropy Rate hypothesis~\citep[CER;][]{genzel-charniak:2002}. To overcome the major difficulty of estimating sentence probabilities conditioned on the previous discourse context, \citet{TingJaeger2012} approximated discourse-level entropy using lexical cues from the previous context. In contrast, we leverage modern computational psycholinguistic neural techniques to obtain word and sentence-level estimates of inter-sentential discourse predictability and study the impact of these measures on Hindi word order choices. 
We conclude that discourse-level priming influences Hindi word order decisions and interpret our findings in the light of the factors outlined by~\citet{reitter2011}.

Hindi (Indo-Aryan language; Indo-European language family) has a rich case-marking system and flexible word order, though it mainly follows SOV word order \citep{kachru2006hindi} as exemplified below.

 \begin{small}
\begin{exe}

  \ex \label{ex:hindi-intro}
  \begin{xlist}
  \ex[]{\label{ex:hindi-order-ref}
    \gll {\underline{amar ujala}-ko} {\bf yah} {\emph{sukravar}-ko} {daak-se} {prapt} {hua}\\ 
         {\underline{Amar Ujala}-\textsc{acc}} {\bf it} {\emph{friday}-on} {post-\textsc{inst}} {receive} {be\textsc{.pst.sg}}\\
     \glt \underline{Amar Ujala} received \textbf{it} by post on \emph{Friday}.}
  \ex[] {\label{ex:hindi-order-var1} {{\bf yah} \underline{{amar ujala}}-ko \emph{sukravar}-ko daak-se prapt hua}} 
  \ex[] {\label{ex:hindi-order-var2} {\emph{sukravar}-ko {\bf yah} \underline{amar ujala}-ko daak-se prapt hua}} 
  \end{xlist}

\end{exe}
\end{small}

To test ordering preferences, we generated meaning-equivalent grammatical variants (Examples~\ref{ex:hindi-order-var1} and \ref{ex:hindi-order-var2} above) of reference sentences (Example~\ref{ex:hindi-order-ref}) from the Hindi-Urdu Treebank corpus of written text~\citep[HUTB;][]{Bhatt2009} by permuting their preverbal constituent ordering. Subsequently, we used a logistic regression model to distinguish the original reference sentences from the plausible variants based on a set of cognitive predictors. We test whether fine-tuning a neural language model on preceding sentences improves predictions of preverbal Hindi constituent order in later sentences over other cognitive control measures. The motivation for our fine-tuning method is that, during reading, encountering a syntactic structure eases the comprehension of subsequent sentences with similar syntactic structures as attested in a wide variety of languages~\citep{arai2007priming,tooley2010syntactic} including Hindi~\citep{husain-yadav2020}. Our cognitive control factors are motivated by recent works which show that Hindi optimizes processing efficiency by minimizing lexical and syntactic surprisal~\citep{ranjan-etal-2019-surprisal} and dependency length \citep{cog:sid} at the sentence level.

Our results indicate that discourse predictability is maximized by reference sentences compared with alternative orderings, indicating that discourse predictability influences Hindi word-order preferences. This finding corroborates previous findings of adaptation/priming in comprehension \citep{fine2013rapid,fine2016role} and production~\citep{gries2005,bock1986syntactic}. Generally, this effect is influenced by lexical priming, but we also find that certain object-fronted constructions prime subsequent object-fronting, providing evidence for self-priming of larger syntactic configurations. With the introduction of neural model surprisal scores, dependency length minimization effects reported to influence Hindi word order choices in previous work~\citep{cog:sid} disappear except in the case of direct object fronting, which we interpret as evidence for the Information Locality Hypothesis~\citep{futrell2020}. Finally, we discuss the implications of our findings for syntactic priming in both comprehension and production.

Our main contribution is that we show the impact of discourse predictability on word order choices using modern computational methods and naturally occurring data (as opposed to carefully controlled stimuli in behavioural experiments). Cross-linguistic evidence is imperative to validate theories of language processing~\citep{Jaeger2009compass}, and in this work we extend existing theories of how humans prioritize word order decisions to Hindi.

\section{Background}\label{sect:backg}


\subsection{Surprisal Theory}\label{subsect:surp}

Surprisal Theory \citep{hale2001,levy2008} posits that comprehenders construct probabilistic interpretations of sentences based on previously encountered structures. Mathematically, the \textit{surprisal} of the ${k}^{th}$ word, $w_k$, is defined as the negative log probability of $w_k$ given the preceding context: 

\begin{small}
\begin{equation}{\label{eq1}}
S_{k}= -\log P(w_{k}|w_{1...k-1}) = \log \frac{P(w_{1}...w_{k-1})}{P(w_{1}...w_{k})}
\end{equation}
\end{small}

These probabilities can be computed either over word sequences or syntactic configurations and reflect the information load (or predictability) of $w_k$. High surprisal is correlated with longer reading times~\citep{levy2008,DembergKeller2008,staub2015} as well as longer spontaneous spoken word durations~\citep{demberg-sayeed-2012,dammalapati2021expectation}. Lexical predictability estimated using \ngram\ language models is one of the strongest determinants of word-order preferences in both English~\citep{cog:raja} and Hindi~\citep{cog:sid,ranjan-etal-2019-surprisal,jain-etal-2018-uniform}.

\subsection{Dependency Locality Theory}

Dependency locality theory \citep{gibson00} has been shown to be effective at predicting the comprehension difficulty of a sequence, with shorter dependencies generally being easier to process than longer ones \citep[][cf.\ \protect\citeauthor{DembergKeller2008}, \citeyear{DembergKeller2008}]{Temperley2007,futrell2015,Liu2017}.

\section{Data and Models}\label{sect:method}

Our dataset comprises 1996 reference sentences containing well-defined subject and object constituents from the HUTB\footnote{\url{https://verbs.colorado.edu
/hindiurdu/}} corpus of dependency trees~\citep{Bhatt2009}. The HUTB corpus, which belongs to the newswire domain and contains written text in a natural discourse context, is a human-annotated, multi-representational, and multi-layered treebank. The dependency trees here assumes Panini’s grammatical model where each sentence is represented as a series of \textit{modifier-modified} elements~\citep{bharati2002anncorra,sangal1995natural}. 
Each tree in the HUTB corpus denotes words in the sentence with nodes such that head words (modified) are linked to the dependent words (modifier) via labelled links denoting the grammatical relationship between word
pairs. 

For each reference sentence in the HUTB corpus, we created artificial variants by permuting the preverbal constituents whose heads were linked to the root node in the dependency tree. Inspired by grammar rules proposed in the NLG literature~\citep{llc:raja:mwhite:2014}, ungrammatical variants were automatically filtered out by detecting dependency relation sequences not attested in the original HUTB corpus. After filtering, we had 72833 variant sentences for our classification task. Figure \ref{fig:hutb-tree} in Appendix~\ref{appendix:A} displays the dependency tree for Example sentence \ref{ex:hindi-order-ref} and explains our variant generation procedure in more detail. 

To determine whether the original word order (i.e.\ the reference sentence) is preferred to the permuted word orders (i.e.\ the variant sentences), we conducted a targeted human evaluation via forced-choice rating task and collected sentence judgments from 12 Hindi native speakers for 167 randomly selected reference-variant pairs in our data set. Participants were first shown the preceding sentence, and then they were asked to select the best continuation between either the reference or the variant. We found that 89.92\% of the reference sentences which originally appeared in the HUTB corpus were also preferred by native speakers compared to the artificially generated grammatical variants expressing similar meaning (Further details are provided in Appendix~\ref{subsec:human-eval}). Therefore, in our analyses we treat the HUTB reference sentences as human-preferred gold orderings compared with other possible automatically-generated constituent orderings.

\subsection{Models}\label{subsect:model}

We set up a binary classification task to separate the original HUTB reference sentences from the variants using the cognitive metrics described in Section \ref{sect:backg}. To alleviate the data imbalance between the two classes (1996 references vs 72833 variants), we transformed our data set using the approach described in \citet{Joachims:2002}. This technique converts a binary classification problem into a pair-wise ranking task by training the classifier on the difference of the feature vectors of each reference and its corresponding variants (see Equations \ref{eq:nor} and \ref{eq:joc}). Equation \ref{eq:nor} displays the objective of a standard binary classifier, where the classifier must learn a feature weight ($w$) such that the dot product of $w$ with the reference feature vector ($\phi(reference)$) is greater than the dot product of $w$ with the variant feature vector ($\phi(variant)$). This objective can be rewritten as equation \ref{eq:joc} such that the dot product of $w$ with the difference of the feature vectors is greater than zero. 

\begin{small}
\begin{equation}\label{eq:nor}
 w~\cdot~\phi(reference) > w~\cdot~\phi(variant)
\end{equation}
\end{small}

\vspace{-1em}

\begin{small}
\begin{equation}\label{eq:joc}
 w~\cdot~(\phi(reference)~-~\phi(variant)) > 0    
\end{equation}
\end{small}

Every variant sentence in our dataset was paired with its corresponding reference sentence with order balanced across these pairings (e.g., Example~\ref{ex:hindi-intro} would yield (\ref{ex:hindi-order-ref},\ref{ex:hindi-order-var1}) and (\ref{ex:hindi-order-var2},\ref{ex:hindi-order-ref})). Thereafter, their feature vectors were subtracted (e.g., \ref{ex:hindi-order-ref}-\ref{ex:hindi-order-var1} and \ref{ex:hindi-order-var2}-\ref{ex:hindi-order-ref}), and binary labels were assigned to each transformed data point. {\it Reference-Variant} pairs were coded as ``1" and {\it Variant-Reference} pairs were coded as ``0". The alternate pair ordering thus re-balanced our previously severely imbalanced classification task. Table \ref{tab:joc-trans} in Appendix \ref{appendix:joc} illustrates the original and transformed values of the independent variables. 

For each reference sentence, our objective was to model the possible syntactic choices entertained by the speaker. In each instance, the author chose to generate the reference order over the variant, implicitly demonstrating an order preference. If the cognitive factors in our study influenced that decision, a logistic regression model should be able to use those factors to predict which syntactic choice was ultimately chosen by the author. Using the transformed features dataset labelled with 1 (denoting a preference for the reference order) and 0 (denoting a preference for the variant order), we trained a logistic regression model to predict each reference sentence (see Equation \ref{eq:regr}). We report our classification results using 10-fold cross-validation. 
The regression results are reported on the entire transformed test data for the respective experiments. All experiments were done with the Generalized Linear Model (GLM) package in $R$.


\vspace{-1em}
\begin{small}
\begin{equation}\label{eq:regr}
choice \sim  \begin{cases}
 & \text{$\delta$ dependency length +} \\ 
 & \text{$\delta$ trigram surp + $\delta$ pcfg surp +} \\ 
 & \text{$\delta$ IS score + $\delta$ lexical repetition surp +}\\ 
 & \text{$\delta$ lstm surp + $\delta$ adaptive lstm surp}
\end{cases}
\end{equation}
\end{small}

Here \textit{choice} is encoded by the binary dependent variable as discussed above ($1$: reference preference and $0$: variant preference). To obtain sentence-level surprisal measures, we summed word-level surprisal of all the words in each sentence. The values for independent variables were calculated as follows.



\begin{enumerate}

  \item {\bf Dependency length}: We computed a sentence-level dependency length measure by summing the head-dependent distances (measured as the number of intervening words) in the HUTB reference and variant dependency trees. 
  
  \item  {\bf Trigram surprisal}: For each word in a sentence, we estimated its local predictability using a 3-gram language model (LM) trained on the written section of the EMILLE Hindi Corpus~\citep{emille2002}, which consists of 1 million mixed genre sentences, using the SRILM toolkit~\citep{SRILM-ICSLP:2002} with Good-Turing discounting.

  \item {\bf PCFG surprisal}: The syntactic predictability of each word in a sentence was estimated using the Berkeley latent-variable PCFG parser\footnote{5-fold cross-validated parser training and testing F1-score metrics were 90.82\% and 84.95\%, respectively.}~\citep{bkp2006}. 
  12000 phrase structure trees were created to train the parser by converting \citeauthor{Bhatt2009}'s HUTB dependency trees into constituency trees using the approach described in \citet{Yadav2017KeepingIS}. Sentence level log-likelihood of each test sentence was estimated by training a PCFG LM on four folds of the phrase structure trees and then testing on a fifth held-out fold.

\item {\bf Information status (IS) score}: We automatically annotated whether each sentence exhibited \textit{given-new} ordering. The subject and object constituents in a sentence were assigned a \textit{Given} tag if its head was a pronoun or any content word within it was mentioned in the preceding sentence. All other phrases were tagged as \textit{New}. For each sentence, IS score was computed as follows: a)~Given-New order = +1 b)~New-Given order = -1 c)~Given-Given and New-New = 0. For an illustration of givenness coding, see Example~\ref{ex:hindi-intro-app} in Appendix \ref{appendix:A} and the description in Appendix \ref{appendix:B}.



  \item \textbf{Lexical repetition surprisal}: For each word in a sentence, we accounted for lexical priming by interpolating a 3-gram language model with a unigram cache LM based on the history of words ($H=100$) containing only the preceding sentence. We used the original implementation provided in the SRILM toolkit with a default interpolation weight parameter ($\mu=0.05$; see Equations \ref{cache1} and \ref{cache2}) based on the approach described by \citet{kuhn1990cache}. The idea is to keep a count of recently occurring words in the sentence history and then boost their probability within the trigram language model. Words that have occurred recently in the text are likely to re-occur\footnote{Out of 13274 sentences present in HUTB, 71.20\% sentences contained at least one content word previously mentioned in the preceding sentence~\cite{jain-etal-2018-uniform}.} in subsequent sentences~\citep{kuhn1990cache,Clarkson97languagemodel}.
  
\begin{scriptsize}
\begin{equation}{\label{cache1}}
\begin{aligned}
P(w_{k}|w_1,w_2,....w_{k-1}) = \mu ~P_{cache}(w_{k}|w_1,w_2,....w_{k-1})\\ + (1-\mu)~{P_{trigram}(w_{k}|w_{k-2}, w_{k-1})}
\end{aligned}
\end{equation}
\begin{equation}{\label{cache2}}
\begin{aligned}
P_{cache}(w_{k}|w_{k-H},w_{k-H+1},..w_{k-1}) = \frac{w_{k}~counts_{(cache)}}{H}
\end{aligned}
\end{equation}
\end{scriptsize}

\item \textbf{LSTM surprisal}: We estimated the predictability for each word according to the entire sentence prefix using a long short-term memory language model~\citep[LSTM;][]{hochreiter1997long} trained on the 1 million written sentences from the EMILLE Hindi corpus~\citep{emille2002}. We used the LSTM implementation provided in the Neural Complexity toolkit~\citep{van2018neural} with default hyper-parameter settings to estimate surprisal using the neural context within each sentence. The exact parameters are as follows: 2 LSTM layers with 200 hidden units each, 200-dimensional word embeddings, 20 units each of learning rate, and training epoch with early stopping. Rest other parameters were set to default setting.\footnote{\url{https://github.com/vansky/neural-complexity\#model-parameters}}

 \item  \textbf{Discourse LSTM surprisal}: We estimated the discourse predictability of each word in the sentence using the {\sc adapt} function of the neural complexity toolkit. \citet{van2018neural} proposed a simple way to continuously adapt a neural LM to each successive test sentence, and found that adaptive surprisal predicts human reading times significantly better than non-adaptive surprisal. Their method takes a pre-trained LSTM LM, and, after generating surprisals for a test sentence, the parameters of the LM get updated based on the cross-entropy loss for that sentence. After that, the revised LM weights are used to predict the next test sentence. This continuous fine-tuning approach effectively modulates a sentence-level LSTM through discourse priming. In our work, for each test sentence, we used our base LSTM LM and adapted it to the immediately preceding context sentence and then used it to generate (discourse-sensitive) surprisal values for the desired sentence. We used an adaptive learning rate of 2 as it minimized the perplexity of the validation data set (see Table~\ref{tab:lr-rate}).%
\footnote{Interestingly, \protect\citet{van2018neural} found that an adaptive learning rate of 2 minimized validation perplexity in English as well, though we leave further investigation of this to future work.}  
\begin{table}[t]
\centering
\scalebox{0.63}{
\begin{tabular}{|l|c|c|c|c|c|c|c|}
\hline
\textbf{Learning Rate} & 0 & 0.002 & 0.02 & 0.2 & \textbf{2} & 20 & 200 \\ \hline
\textbf{Perplexity} & 103.29 & 98.79 & 87.78 & 66.64 & \textbf{56.86} & 117.91 & $\sim10^9$ \\ \hline 
\end{tabular}}
\caption{Learning rate influence on adaptive LSTM validation perplexity ($N = 13274$ sentences; the initial non-adaptive model uses a learning rate of 0)}
\label{tab:lr-rate}
\end{table}

\end{enumerate}

\section{Experiments and Results}\label{sect:results}

We tested the hypothesis that discourse predictability (estimated from adaptive LSTM and lexical repetition surprisal) influences constituent ordering in Hindi over other baseline cognitive controls, including dependency length, information status and trigram and non-adaptive surprisal. The adaptive LSTM surprisal had a high correlation with all other surprisal features and a low correlation with dependency length and information status score (see Figure \ref{fig:corr-plot} in Appendix~\ref{appendix:C}). We report the results of regression and prediction experiments on the full data set as well as on subsets of the data consisting of two types of non-canonical constructions.

\subsection{Regression Analysis}\label{lex:reg}

Our regression results over the entire data set (Table \ref{tab:regr-results1}) show that all of our measures are significant predictors for the task of classifying reference and variant sentences. The negative regression coefficients for our surprisal metrics (including adaptive LSTM surprisal) indicate that surprisal is consistently lower in the reference sentences compared with the competing variants. And adding adaptive discourse LSTM surprisal into a model containing all other predictors significantly improved the fit of our regression model ($\chi^2$ = 66.81; p $<$ 0.001). Thus these results support our core hypothesis that word order choices seem to maximize discourse predictability compared with possible alternative productions. The positive regression coefficient for information status (IS) score indicates that reference sentences adhere to \textit{given-new} ordering. Similarly, adding IS score into a model containing all other predictors significantly improved the fit of our regression model ($\chi^2$ = 127.94; p $<$ 0.001). However, the positive regression coefficient of dependency length suggests that reference sentences exhibit \emph{longer} dependency lengths compared to their variant counterparts, violating locality considerations. Thus dependency length might be in conflict with (and/or overridden by) other factors like discourse priming or information locality (see Section~\ref{sect:disc} for more discussion of this idea). 


\begin{table}
\centering
\begin{small}
\scalebox{1.1}{
\begin{tabular}{lccc}
\toprule
Predictor & $\hat\beta$ & $\hat\sigma$ & t\tabularnewline
\midrule 
intercept  & 1.50  & 0.001  & 1496.47 \\
trigram surprisal  & -0.08  & 0.005  & -14.53 \\
dependency length  & 0.02  & 0.001  & 15.55 \\
pcfg surprisal  & -0.07  & 0.002  & -39.46 \\
IS score  & 0.01  & 0.001  & 11.32 \\
lex-rept surprisal  & -0.03  & 0.005  & -5.31 \\
lstm surprisal  & -0.14  & 0.016  & -9.26 \\
adaptive lstm surprisal  & -0.13  & 0.016  & -8.18\\
\bottomrule 
\end{tabular}}
\caption{Regression model on full data set ($N=72833$; all significant predictors denoted by $|$t$|$\textgreater{}2)}
\label{tab:regr-results1}
\end{small}
\end{table}

We also examined the contribution of each predictor on two non-canonical syntactic configurations, \emph{direct object (DO) fronted} and \emph{indirect object (IO) fronted} constructions, which have been studied extensively in the sentence comprehension literature. Prior work has shown that salient objects tend to occur early in the sentence, thus leading to fronting~\citep{wierzba2020factors,kaiser2004role}. In the specific context of Hindi, \citet{vasishthysall04} examined the role of locality effects in processing these non-canonical word orders in salient as well as non-salient contexts. He showed that the increased distance to the verb in DO-fronted sentences leads to high self-paced reading times at the inner-most verb as compared to its canonical counterpart in both salient and non-salient conditions. However, in IO-fronted constructions, he found that salient contexts alleviated the processing difficulty which was caused by increased distance. Based on these findings, we predict that \textbf{adaptive surprisal should be more effective in IO-fronted than DO-fronted constructions}.



To test this hypothesis, we isolated reference sentences where the direct object precedes the subject (for a DO-fronted test set) and reference sentences where the indirect object precedes the subject (for an IO-fronted test set) along with their context sentences. We compared both sets to paired variants that exhibited canonical order (i.e.\ where the subject preceded both objects). Tables~\ref{tab:lexadpt-do-regres} and \ref{tab:lexadpt-io-regres} present regression results for DO- and IO-fronted constructions respectively. These subsets constitute a very small fraction of our dataset due to the infrequency of these constructions in Hindi. The regression coefficient for adaptive LSTM surprisal was significantly negative for both subsets, indicating that the non-canonical structures are more common in the context of similarly non-canonical structures. This pattern is more robust for IO-fronted reference sentences ($\chi^2$ = 90.90; p $<$ 0.001) than for DO-fronted reference sentences ($\chi^2$ = 4.03; p $=$ 0.04), validating our proposed prediction about these constructions. Coming to the efficacy of IS scores over these two non-canonical constructions, \textit{givenness} is effective in case of DO-fronted reference sentences only ($\chi^2$ = 49.06; p $<$ 0.001). 
Furthermore, in contrast to the IO-fronted subset, the regression coefficient for dependency length in DO-fronted items is significantly negative suggesting that locality considerations are limited to constructions involving a high dependency length difference between reference and variants,\footnote{The average dependency length difference for the DO-subset is 13.92 and for the IO-subset is 7.77 words.} a similar finding to that reported in \citet{cog:sid} on the same task.

\begin{table*}
\begin{center}
\scalebox{1.1}{
\centering
\subfloat[Subtable 1 list of tables text][{Direct objects (DO; 1663 points) fronted cases}]
{
\begin{small}
\begin{tabular}{lrrr}
\toprule
Predictor & $\hat\beta$ & $\hat\sigma$ & t\tabularnewline
\midrule
intercept & \textbf{1.49} & 0.008 & 171.18 \\
trigram surp & \textbf{-0.28} & 0.049 & -5.84 \\
dep length & \textbf{-0.05} & 0.008 & -6.22 \\
pcfg surp & 0.001 & 0.014 & 0.12 \\
IS score & \textbf{0.04} & 0.006 & 7.04 \\
lex repetition surp & 0.07 & 0.044 & 1.67 \\
lstm surp & 0.03 & 0.114 & 0.23\\ 
adaptive lstm surp & \textbf{-0.23} & 0.113 & -2.00 \\
\bottomrule
\end{tabular}
\label{tab:lexadpt-do-regres}
\end{small}
}
\quad\quad
\subfloat[Subtable 2 list of tables text][{Indirect objects (IO; 1353 points) fronted cases}]
{
\begin{small}
\begin{tabular}{lrrr}
\toprule
Predictor & $\hat\beta$ & $\hat\sigma$ & t\tabularnewline
\midrule 
intercept & \textbf{1.51} & 0.008 & 188.49 \\
trigram surp & \textbf{-0.18} & 0.039 & -4.54\\
dep length & 0.02 & 0.012 & 1.77\\
pcfg surp & \textbf{-0.13} & 0.015 & -8.34\\
IS score & -0.01 & 0.005 & -1.87\\
lex repetition surp & 0.03 & 0.036 & 0.92\\
lstm surp & \textbf{1.21} & 0.154 & 7.87\\
adaptive lstm surp & \textbf{-1.50} & 0.155 & -9.67\\
\bottomrule 
\end{tabular}
\label{tab:lexadpt-io-regres}
\end{small}
}}
\caption{Discourse adaptation regression model on DO/IO fronted cases (all significant predictors denoted by $|$t$|$\textgreater{}2)}
\label{tab:exp3-reg}
\end{center}
\end{table*}

\subsection{Prediction Accuracy}\label{lex:class}

\begin{table*}
\centering
\scalebox{0.75}{
\begin{tabular}{|l|c|c|c|c|c|c|c|c|}
\hline
\textbf{Predictors} & \textbf{\begin{tabular}[c]{@{}c@{}}Full\\ Accuracy \%\end{tabular}} & \textbf{DO} & \textbf{IO} & \multirow{8}{*}{} & \textbf{Predictors} & \textbf{\begin{tabular}[c]{@{}c@{}}Full\\ Accuracy \%\end{tabular}} & \textbf{DO} & \textbf{IO} \\ \cline{1-4} \cline{6-9} 
a = IS score & 51.84 & 53.88 & 50.92 &  & \multicolumn{4}{c|}{Collective: with repetition effects} \\ \cline{1-4} \cline{6-9} 
b = dep length & 62.31*** & 68.49*** & 58.91*** &  & base1 = a+b+c+d+e+f & \textbf{95.05} & 80.99 & 89.06 \\ \cline{1-4} \cline{6-9} 
c = pcfg surp & 86.86*** & 65.90 & 78.86*** &  & base1 + g & 95.06 & 81.06 & \textbf{89.65*} \\ \cline{1-4} \cline{6-9} 
d = lex repetition surp & 90.07*** & 77.33*** & 85.07*** &  & \multicolumn{4}{c|}{\multirow{2}{*}{Collective: without repetition effects}} \\ \cline{1-4}
e = 3-gram surp & 91.18*** & 78.95* & 87.29** &  & \multicolumn{4}{c|}{} \\ \cline{1-4} \cline{6-9} 
f = lstm surp & \textbf{94.01***} & 79.55 & 87.28 &  & base2 = a+b+c+e+f & 95.06 & 81.24 & 89.65 \\ \cline{1-4} \cline{6-9} 
g = adaptive lstm surp & 94.06 & \textbf{79.97} & \textbf{88.32***} &  & base2 + g & \textbf{95.09*} & 81.42 & 89.80 \\ \hline
\end{tabular}}
\caption{Prediction performances (Full data set (72833 points), Direct objects (DO; 1663 points) and indirect object (IO; 1353 points) fronted cases; each row refers to a distinct model; *** McNemar's two-tailed significance compared to model on previous row)}
\label{tab:lex-adapt-pred-acc}
\end{table*}

While the previous section explored how predictors contribute to Hindi ordering preferences across all of the data in aggregate, in this section we frame our model as a classification task on held-out data to determine how many sentences are affected by each predictor. This enables us to examine the relative performance of different predictors in identifying Hindi reference sentences amidst artificially generated grammatical variants and to conduct more detailed error analysis of our results. We used 10-fold cross-validation to evaluate model classification accuracy, i.e.\ the percentage of data points where a model correctly predicted the referent sentence over a paired variant, for different subsets of predictors (see Table \ref{tab:lex-adapt-pred-acc}). 

Non-adaptive LSTM surprisal (94.01\% accuracy) and adaptive LSTM surprisal (94.06\%) yielded the best classification accuracies when no other predictors were included. Over a baseline model comprised of every other feature except lexical repetition surprisal (see \emph{base2} in Table \ref{tab:lex-adapt-pred-acc}), adaptive LSTM surprisal induced a small but significant increase of 0.03\% in accuracy (p = 0.04 using McNemar's two-tailed test). When we included lexical repetition surprisal in the baseline model (see \emph{base1} in Table \ref{tab:lex-adapt-pred-acc}), adaptive LSTM surprisal ceased to be a significant predictor. This suggests that, in the general case, the maximization of discourse predictability is driven by localized lexical priming captured by our trigram cache model. Apart from the content words, adaptive LSTM surprisal accounts for the re-occurrence of function words (e.g., case markers) which have been shown to modulate syntactic priming and drive parsing processes~\citep{husain-yadav2020}.


To study prediction accuracy on non-canonical constructions, we restricted our analyses to IO- and DO-fronted items in the test partition (still training the classifier on the full training partition for each fold). In contrast to the DO-fronted subset, adaptive surprisal was a significant predictor of IO-fronted syntactic choice, even in the presence of lexical repetition surprisal, as is evident from the significant increase of 0.6\% in accuracy (p = 0.02 using McNemar's two-tailed test; see the rightmost IO column in Table \ref{tab:lex-adapt-pred-acc}). This result indicates that discourse predictability is effective in predicting IO-fronting in sentences that follow other IO-fronted sentences, suggesting the presence of syntactic priming effects. {We consider adaptive LSTM LM surprisal (i.e., updating LM weights on successive sentences at test time) as an indicative of syntactic priming in this work but not the vanilla LSTM LM surprisal.} We present a more nuanced discussion on this theme in Section \ref{sect:disc}.

Both our regression and classification results demonstrate that discourse adaptation is more effective in IO-fronted than DO-fronted constructions, mirroring the findings in Hindi sentence comprehension, where \citet{vasishthysall04} showed that discourse context could compensate for the processing difficulty induced by indirect object fronting. The findings of our computational modelling reported in Table \ref{tab:lex-adapt-pred-acc} are further validated by the agreement accuracy of our human evaluation study described in Section~\ref{sect:method}. Participants were more prone to prefer IO-fronted construction (80\%) compared to DO-fronted construction (65\%) as shown in Table~\ref{tab:human-eval} of Appendix~\ref{subsec:human-eval}. 

\subsection{Qualitative Analysis: Success of Adaptive LSTM Surprisal}


Further linguistic analyses in IO-fronted constructions revealed that LSTM adaptation also captured the priming of \textit{given-given} items, potentially modeling the preferred ordering of multiple \emph{given} items, a case not captured by IS score or lexical repetition surprisal. Reference sentence \ref{ex:hindi-order-ref} is correctly predicted by the model containing adaptive LSTM surprisal and all other features (i.e., \textit{base1+g} in Table \ref{tab:lex-adapt-pred-acc}) but a model without adaptive LSTM surprisal (i.e., base1) predicts the variant Example~\ref{ex:hindi-order-var1}. Appendix~\ref{appendix:io-fronting} Table \ref{tab:sent-level-scores} presents the exact scores of different predictors for the referent-variant pairs (\ref{ex:hindi-order-ref} and \ref{ex:hindi-order-var1}). All predictors but LSTM and adaptive LSTM surprisal assign high score for the reference sentence with respect to its paired variant. Adaptive LSTM surprisal assigns a low per-word surprisal for the phrase {\it amar ujala} when it comes at the first position in the reference sentence (\ref{ex:hindi-order-ref}) with respect to when it comes at the second position in the variant (\ref{ex:hindi-order-var1}), potentially modeling \textit{givenness} as this word occurred in the previous sentence (Example~\ref{ex:hindi-prev-sent-app} in Appendix~\ref{appendix:io-fronting}) as well. See Figure \ref{fig:io-profile} in Appendix~\ref{appendix:io-fronting} for the information profile of the reference-variant pairs.

\subsection{What causes priming?}

In the priming literature, there is debate as to whether priming is driven by residual neural activation (short-lived effects) or by humans learning and updating their language expectations (long-lived effects). \citet{BockGriff2000} showed that syntactic priming in humans persisted even when prime and target sentences were separated by 10 intervening sentences, supporting the implicit learning (long-lived) hypothesis of syntactic priming. In order to test this effect on constituent ordering choice, we repeated our adaptation experiment by adapting to additional context sentences from the preceding discourse. Adaptive LSTM surprisal and lexical repetition surprisal were estimated by adapting the base LSTM LM and trigram LM, respectively, to five preceding context sentences, rather than the single context sentence we used for our other analyses. We found that for non-canonical IO/DO-fronted constructions, additional context sentences do not improve the adaptive LSTM LM's word order predictions, suggesting that priming may be driven by short-term residual activation (see Table {\ref{tab:3vs5:pred}} in the Appendix \ref{appendix:prev5}).

\section{Variance Inflation Factor}

In this section, we evaluate our regression models for multicollinearity in terms of variance inflation factor (VIF) score. 
As Figure \ref{fig:corr-plot} in Appendix \ref{appendix:C} denotes, the adaptive LSTM surprisal measure has a high correlation with all other surprisal predictors, which raises some suspicion that estimates of effects of the variables in our regression model might be unreliable. 
Table \ref{tab:vif-score} in Appendix \ref{subsec:vif} presents the VIF scores for different regression models containing different set of predictors on full dataset, DO- and IO-fronted subsets. The outcomes indicate that all the surprisal predictors except PCFG surprisal in our original regression models have very high VIF scores (see Table \ref{tab:orig-vif} in Appendix \ref{subsec:vif}). Nevertheless, removing the more correlated measures, such as trigram surprisal and base LSTM surprisal does not alter any of our old results (see Table \ref{tab:regr-results-vif} for new regression results and Table \ref{tab:trans-vif} for corresponding VIF scores in Appendix \ref{subsec:vif}).
All the surprisal measures including adaptive LSTM and lexical repetitions surprisal have negative regression coefficients suggesting that Hindi tends to optimize for discourse predictability.
\section{Discussion}\label{sect:disc}

Our main findings suggest that in written Hindi, people choose word orders that maximize discourse predictability. The actual psychological mechanisms are conceivably lexical and structural priming. Our results indicate that lexical priming is most influential in canonical sentence contexts, but syntactic priming does influence ordering preferences in non-canonical contexts. Below, we discuss the implications of our findings in terms of the 4 factors affecting syntactic priming discussed in detail by \citet{reitter2011}: \textit{inverse frequency interaction, decay, lexical boost, and cumulativity}. The IO-fronted construction is very rare (0.76\% of our data) compared to DO-fronted non-canonical sentences (1\% of our data) in the HUTB corpus of 13274 sentences. We find strong priming effects in IO-fronted constructions but weak priming in the case of DO-fronted constructions, providing evidence for an \textit{inverse frequency interaction}~\citep{Scheepers2003,JaegerSnider2007}.

%
Our finding that priming is not aided by long-term contexts indicates a \textit{decay effect} in priming, which supports the residual activation (short-lived) hypothesis of priming in comprehension~\citep{Pickering1998}. Nevertheless, there has been evidence for implicit learning effects in comprehension as well~\citep{luka2005structural,wells2009experience}. 
More recently, \citet{ranjan2022dmph} using a similar setup as our current work argued for the existence of both the accounts \emph{viz.,} residual activation and implicit learning, and demonstrated the role of dual mechanism priming effects~\citep{tooley2010syntactic} in Hindi word order.

Previous work suggests that lexical overlap between prime and target sentences enhances syntactic priming~\citep{Pickering1998,gries2005}. The repeated lexical items become cues during sentence planning and bias the speaker to produce similar structures that those repeated lexical items tend to occur in. Overall, we find that lexical repetition influences Hindi syntactic choice; however, syntactic priming is observed over and above lexical repetition in non-canonical constructions. It's interesting to note that comparable results have also been reported for English dialogue corpora~\citep{healey2014divergence,green2021global}. 
We plan to conduct a systematic investigation on Hindi spoken data as a part of future work. 


Finally, with regards to the \textit{cumulativity} of priming, \citet{JaegerSnider2007} showed in their corpus study of production of passives and \textit{that}-insertion/omission that the effect of priming increases with the number of primes preceding it. Our work does not investigate this specifically, and more controlled experiments would be required. 

The success of LSTM-based surprisal estimates over and above dependency length can also be interpreted in light of Futrell's \citeyearpar{futrell-2019-information} point about the limitation of Surprisal Theory with respect to word order. Futrell modified Surprisal Theory by positing that the per-word processing difficulty is proportional to its surprisal given a {\em lossy memory representation} of the preceding context. Moreover,~\citet{futrell2020} proposed the Information Locality Hypothesis (ILH) which states that all pairs of words with high mutual information (not merely syntactically related words) tend to be located close to one another. The long window offered by LSTM surprisal thus models relationships between words at varying distances (over and above conventional trigram models). The success of these surprisal estimates for the task of reference sentence prediction provides some preliminary evidence for ILH in the case of word order. 

Future work needs to tease apart priming effects of both vanilla LSTM and adaptive LSTM surprisal in the light of recent works. In this work, sentences are treated as independent while estimating their surprisal using vanilla LSTM LM, so there is no way vanilla LSTM can exhibit syntactic priming given the independent sentences.
However,~\citet{misra2020exploring} demonstrated that BERT exhibits ``priming effect''. The BERT LM was able to predict a word with greater probability when the context included a related word than an unrelated word. However, the effect decreased as the amount of information provided by the context increases. In other words, the related prime under high contextual constraint started acting as distractor---actively demoting the target word in the probability distribution; thus exhibiting ``mispriming effect''~\cite{kassner-schutze-2020-negated}. This could be due to stylistic avoidance of repeated structures/words in the adjacent sentences. 
Future work also needs to investigate whether word-order preferences can be jointly optimized using multiple factors~\citep{GildeaJaeger2015}. In particular, the relationship between the drive to minimize surprisal (as found in this work) and the tendency to make information profiles uniform~\citep{Jaeger:2010cogpsych} needs to be explored more thoroughly in the light of recent findings~\citep{meister-etal-2021-revisiting}.

Overall, our results demonstrate that Hindi word order preferences are influenced by discourse predictability maximization considerations. The actual mechanisms of discourse effects are plausibly lexical and syntactic priming.



\section{Limitations}

The `levels' problem discussed in~\citet{levy2018} which posits 2 levels of linguistic optimisation is germane while evaluating our work. Our results are restricted to the level of syntactic choices made by individual speakers or users of a given language over a lifetime (and not at the level of entire grammars and evolutionary timescales). Our experiments conducted on written text need to be performed on spoken data in order to make claims about priming in language production.

\section*{Acknowledgements}

We thank the first author's dissertation committee members, Drs. Mausam and Samar Husain, as well as the Cornell C.Psyd members for their insightful comments and suggestions on this work. We thank Rupesh Pandey's logistical assistance in gathering the human judgment data for this work. Additionally, we are grateful for the thorough feedback provided by the anonymous reviewers of EMNLP 2022, ACL ARR 2021, and COMCO 2021. The first author acknowledges the extramural travel grant from Microsoft Research India to present the work at EMNLP 2022. Finally, the last two authors also thank extramural funding from the Department of Science and Technology of India through the Cognitive Science Research Initiative (project no. DST/CSRI/2018/263).

\bibliographystyle{aclnatbib}
\bibliography{bibfile,extra}

\begin{thebibliography}{67}
\expandafter\ifx\csname natexlab\endcsname\relax\def\natexlab#1{#1}\fi

\bibitem[{Arai et~al.(2007)Arai, Van~Gompel, and Scheepers}]{arai2007priming}
Manabu Arai, Roger~PG Van~Gompel, and Christoph Scheepers. 2007.
\newblock Priming ditransitive structures in comprehension.
\newblock \emph{Cognitive psychology}, 54(3):218--250.

\bibitem[{Baker et~al.(2002)Baker, Hardie, McEnery, Cunningham, and
  Gaizauskas}]{emille2002}
Paul Baker, Andrew Hardie, Tony McEnery, Hamish Cunningham, and Robert
  Gaizauskas. 2002.
\newblock Emille: a 67-million word corpus of indic languages: data collection,
  mark-up and harmonization.
\newblock In \emph{Proceedings of LREC 2002}, pages 819--827. Lancaster
  University.

\bibitem[{Bharati et~al.(2002)Bharati, Sangal, Chaitanya, Kulkarni, Sharma, and
  Ramakrishnamacharyulu}]{bharati2002anncorra}
Akshar Bharati, Rajeev Sangal, Vineet Chaitanya, Amba Kulkarni, Dipti~Misra
  Sharma, and KV~Ramakrishnamacharyulu. 2002.
\newblock Anncorra: building tree-banks in indian languages.
\newblock In \emph{COLING-02: The 3rd Workshop on Asian Language Resources and
  International Standardization}.

\bibitem[{Bhatt et~al.(2009)Bhatt, Narasimhan, Palmer, Rambow, Sharma, and
  Xia}]{Bhatt2009}
Rajesh Bhatt, Bhuvana Narasimhan, Martha Palmer, Owen Rambow, Dipti~Misra
  Sharma, and Fei Xia. 2009.
\newblock \href {http://dl.acm.org/citation.cfm?id=1698381.1698417} {A
  multi-representational and multi-layered treebank for {H}indi/urdu}.
\newblock In \emph{Proceedings of the Third Linguistic Annotation Workshop},
  ACL-IJCNLP '09, pages 186--189, Stroudsburg, PA, USA. Association for
  Computational Linguistics.

\bibitem[{Bock and Griffin(2000)}]{BockGriff2000}
Bock and Z.~Griffin. 2000.
\newblock The persistence of structural priming: Transient activation or
  implicit learning.
\newblock \emph{Journal of Experimental Psychology}, 2(120):177--192.

\bibitem[{Bock(1986)}]{bock1986syntactic}
J.Kathryn Bock. 1986.
\newblock \href {https://doi.org/https://doi.org/10.1016/0010-0285(86)90004-6}
  {Syntactic persistence in language production}.
\newblock \emph{Cognitive Psychology}, 18(3):355 -- 387.

\bibitem[{Christiansen and Chater(2008)}]{christiansen-chater-2008}
Morten~H. Christiansen and Nick Chater. 2008.
\newblock \href {https://doi.org/10.1017/S0140525X08004998} {Language as shaped
  by the brain}.
\newblock \emph{Behavioral and Brain Sciences}, 31(5):489–509.

\bibitem[{Clarkson and Robinson(1997)}]{Clarkson97languagemodel}
P.R. Clarkson and A.~J. Robinson. 1997.
\newblock Language model adaptation using mixtures and an exponentially
  decaying cache.
\newblock In \emph{In Proceedings of ICASSP-97}, pages 799--802.

\bibitem[{Dammalapati et~al.(2021)Dammalapati, Rajkumar, and
  Agarwal}]{dammalapati2021expectation}
Samvit Dammalapati, Rajakrishnan Rajkumar, and Sumeet Agarwal. 2021.
\newblock \href{https://doi.org/10.7275/s5yn-7d90}{Effects of Duration,
  Locality, and Surprisal in Speech Disfluency Prediction in English
  Spontaneous Speech}.
\newblock In \emph{Proceedings of the Society for Computation in Linguistics},
  volume~4, page~10.

\bibitem[{Demberg and Keller(2008)}]{DembergKeller2008}
Vera Demberg and Frank Keller. 2008.
\newblock \href
  {http://scholar.google.com/scholar.bib?q=info:1ulLoWI1IDoJ:scholar.google.com/&output=citation&hl=de&as_sdt=0,5&ct=citation&cd=0}
  {Data from eye-tracking corpora as evidence for theories of syntactic
  processing complexity}.
\newblock \emph{Cognition}, 109(2):193--210.

\bibitem[{Demberg et~al.(2012)Demberg, Sayeed, Gorinski, and
  Engonopoulos}]{demberg-sayeed-2012}
Vera Demberg, Asad~B. Sayeed, Philip~J. Gorinski, and Nikolaos Engonopoulos.
  2012.
\newblock \href {http://dl.acm.org/citation.cfm?id=2390948.2390992} {Syntactic
  surprisal affects spoken word duration in conversational contexts}.
\newblock In \emph{Proceedings of the 2012 Joint Conference on Empirical
  Methods in Natural Language Processing and Computational Natural Language
  Learning}, EMNLP-CoNLL '12, pages 356--367, Stroudsburg, PA, USA. Association
  for Computational Linguistics.

\bibitem[{Fine and Jaeger(2016)}]{fine2016role}
Alex~B Fine and T~Florian Jaeger. 2016.
\newblock The role of verb repetition in cumulative structural priming in
  comprehension.
\newblock \emph{Journal of Experimental Psychology: Learning, Memory, and
  Cognition}, 42(9):1362.

\bibitem[{Fine et~al.(2013)Fine, Jaeger, Farmer, and Qian}]{fine2013rapid}
Alex~B Fine, T~Florian Jaeger, Thomas~A Farmer, and Ting Qian. 2013.
\newblock Rapid expectation adaptation during syntactic comprehension.
\newblock \emph{PloS one}, 8(10):e77661.

\bibitem[{Futrell(2019)}]{futrell-2019-information}
Richard Futrell. 2019.
\newblock \href {https://doi.org/10.18653/v1/W19-7902} {Information-theoretic
  locality properties of natural language}.
\newblock In \emph{Proceedings of the First Workshop on Quantitative Syntax
  (Quasy, SyntaxFest 2019)}, pages 2--15, Paris, France. Association for
  Computational Linguistics.

\bibitem[{Futrell et~al.(2020)Futrell, Gibson, and Levy}]{futrell2020}
Richard Futrell, Edward Gibson, and Roger~P. Levy. 2020.
\newblock \href {https://doi.org/10.1111/cogs.12814} {Lossy-context surprisal:
  An information-theoretic model of memory effects in sentence processing}.
\newblock \emph{Cognitive Science}, 44(3):e12814.

\bibitem[{Futrell et~al.(2015)Futrell, Mahowald, and Gibson}]{futrell2015}
Richard Futrell, Kyle Mahowald, and Edward Gibson. 2015.
\newblock \href {https://doi.org/10.1073/pnas.1502134112} {Large-scale evidence
  of dependency length minimization in 37 languages}.
\newblock \emph{Proceedings of the National Academy of Sciences},
  112(33):10336--10341.

\bibitem[{Genzel and Charniak(2002)}]{genzel-charniak:2002}
Dmitriy Genzel and Eugene Charniak. 2002.
\newblock \href {https://doi.org/10.3115/1073083.1073117} {Entropy rate
  constancy in text}.
\newblock In \emph{Proceedings of the 40th Annual Meeting on Association for
  Computational Linguistics}, ACL '02, pages 199--206, Stroudsburg, PA, USA.
  Association for Computational Linguistics.

\bibitem[{Gibson(2000)}]{gibson00}
Edward Gibson. 2000.
\newblock \href
  {http://www.ling.uni-potsdam.de/~vasishth/Papers/Gibson-Cognition2000.pdf}
  {Dependency locality theory: {A} distance-based theory of linguistic
  complexity}.
\newblock In Alec Marantz, Yasushi Miyashita, and Wayne O'Neil, editors,
  \emph{Image, Language, brain: {P}apers from the First Mind Articulation
  Project Symposium}. MIT Press, Cambridge, MA.

\bibitem[{Gibson et~al.(2019)Gibson, Futrell, Piandadosi, Dautriche, Mahowald,
  Bergen, and Levy}]{Gibson2019}
Edward Gibson, Richard Futrell, Steven~T. Piandadosi, Isabelle Dautriche, Kyle
  Mahowald, Leon Bergen, and Roger Levy. 2019.
\newblock \href {https://doi.org/10.1016/j.tics.2019.02.003} {How efficiency
  shapes human language}.
\newblock \emph{Trends in Cognitive Sciences}, 23(5):389--407.

\bibitem[{Gildea and Jaeger(2015)}]{GildeaJaeger2015}
Daniel Gildea and T.~Florian Jaeger. 2015.
\newblock \href {http://arxiv.org/abs/1510.02823} {Human languages order
  information efficiently}.
\newblock \emph{CoRR}, abs/1510.02823.

\bibitem[{Green and Sun(2021)}]{green2021global}
Clarence Green and He~Sun. 2021.
\newblock \href {https://doi.org/10.1016/j.langsci.2020.101353} {Global
  estimates of syntactic alignment in adult and child utterances during
  interaction: Nlp estimates based on multiple corpora}.
\newblock \emph{Language Sciences}, 85:101353.

\bibitem[{Gries(2005)}]{gries2005}
Stefan~Th. Gries. 2005.
\newblock \href {https://doi.org/10.1007/s10936-005-6139-3} {Syntactic priming:
  A corpus-based approach}.
\newblock \emph{Journal of Psycholinguistic Research}, 34(4):365--399.

\bibitem[{Hale(2001)}]{hale2001}
John Hale. 2001.
\newblock \href {https://doi.org/10.3115/1073336.1073357} {A probabilistic
  {Earley} parser as a psycholinguistic model}.
\newblock In \emph{Proceedings of the second meeting of the North American
  Chapter of the Association for Computational Linguistics on Language
  technologies}, NAACL '01, pages 1--8, Pittsburgh, Pennsylvania. Association
  for Computational Linguistics.

\bibitem[{Hawkins(1994)}]{hawkins1994}
John~A. Hawkins. 1994.
\newblock \emph{A {P}erformance Theory of Order and Constituency}.
\newblock Cambridge University Press, New York.

\bibitem[{Hawkins(2000)}]{Hawkins2000}
John~A. Hawkins. 2000.
\newblock \href {https://doi.org/10.1017/S0954394599113012} {The relative order
  of prepositional phrases in english: Going beyond manner-place-time}.
\newblock \emph{Language Variation and Change}, 11(03):231--266.

\bibitem[{Hawkins(2004)}]{hawkins04}
John~A. Hawkins. 2004.
\newblock \emph{Efficiency and Complexity in Grammars}.
\newblock Oxford University Press.

\bibitem[{Hawkins(2014)}]{Hawkins2014}
John~A. Hawkins. 2014.
\newblock \emph{Cross-Linguistic Variation and Efficiency}.
\newblock Oxford University Press.

\bibitem[{Healey et~al.(2014)Healey, Purver, and Howes}]{healey2014divergence}
Patrick~GT Healey, Matthew Purver, and Christine Howes. 2014.
\newblock \href {https://doi.org/10.1371/journal.pone.0098598} {Divergence in
  dialogue}.
\newblock \emph{PloS one}, 9(6):e98598.

\bibitem[{Hochreiter and Schmidhuber(1997)}]{hochreiter1997long}
Sepp Hochreiter and J{\"u}rgen Schmidhuber. 1997.
\newblock Long short-term memory.
\newblock \emph{Neural computation}, 9(8):1735--1780.

\bibitem[{Husain and Yadav(2020)}]{husain-yadav2020}
Samar Husain and Himanshu Yadav. 2020.
\newblock \href {https://doi.org/10.3389/fpsyg.2020.00454} {Target complexity
  modulates syntactic priming during comprehension}.
\newblock \emph{Frontiers in Psychology}, 11:454.

\bibitem[{Jaeger(2010)}]{Jaeger:2010cogpsych}
T.~Florian Jaeger. 2010.
\newblock \href {http://dx.doi.org/10.1016/j.cogpsych.2010.02.002} {Redundancy
  and reduction: Speakers manage information density}.
\newblock \emph{Cognitive Psychology}, 61(1):23--62.

\bibitem[{Jaeger and Norcliffe(2009)}]{Jaeger2009compass}
T.~Florian Jaeger and Elizabeth Norcliffe. 2009.
\newblock \href {http://dx.doi.org/10.1111/j.1749-818X.2009.00147.x} {The
  cross-linguistic study of sentence production: State of the art and a call
  for action}.
\newblock \emph{Language and Linguistic Compass}, 3(4):866--887.

\bibitem[{Jaeger and Snider(2007)}]{JaegerSnider2007}
T.~Florian Jaeger and Neal Snider. 2007.
\newblock Implicit learning and syntactic persistence: Surprisal and
  cumulativity.
\newblock \emph{University of Rochester Working Papers in the Language
  Sciences}, 3:26--44.

\bibitem[{Jaeger and Tily(2011)}]{JaegerTily2011}
T.~Florian Jaeger and Harold Tily. 2011.
\newblock \href {https://doi.org/10.1002/wcs.126} {Language processing
  complexity and communicative efficiency}.
\newblock \emph{WIRE: Cognitive Science}, 2(3):323--335.

\bibitem[{Jain et~al.(2018)Jain, Singh, Ranjan, Rajkumar, and
  Agarwal}]{jain-etal-2018-uniform}
Ayush Jain, Vishal Singh, Sidharth Ranjan, Rajakrishnan Rajkumar, and Sumeet
  Agarwal. 2018.
\newblock \href {https://www.aclweb.org/anthology/W18-4605}
  {\href{https://www.aclweb.org/anthology/W18-4605/}{Uniform Information
  Density Effects on Syntactic Choice in {H}indi}}.
\newblock In \emph{Proceedings of the Workshop on Linguistic Complexity and
  Natural Language Processing}, pages 38--48, Santa Fe, New-Mexico. Association
  for Computational Linguistics.

\bibitem[{Joachims(2002)}]{Joachims:2002}
Thorsten Joachims. 2002.
\newblock \href {https://doi.org/10.1145/775047.775067} {Optimizing search
  engines using clickthrough data}.
\newblock In \emph{Proceedings of the Eighth ACM SIGKDD International
  Conference on Knowledge Discovery and Data Mining}, KDD '02, pages 133--142,
  New York, NY, USA. ACM.

\bibitem[{Kachru(2006)}]{kachru2006hindi}
Y.~Kachru. 2006.
\newblock \emph{{H}indi}.
\newblock London Oriental and African language library. John Benjamins
  Publishing Company.

\bibitem[{Kaiser and Trueswell(2004)}]{kaiser2004role}
Elsi Kaiser and John~C Trueswell. 2004.
\newblock The role of discourse context in the processing of a flexible
  word-order language.
\newblock \emph{Cognition}, 94(2):113--147.

\bibitem[{Kassner and Sch{\"u}tze(2020)}]{kassner-schutze-2020-negated}
Nora Kassner and Hinrich Sch{\"u}tze. 2020.
\newblock \href {https://doi.org/10.18653/v1/2020.acl-main.698} {Negated and
  misprimed probes for pretrained language models: Birds can talk, but cannot
  fly}.
\newblock In \emph{Proceedings of the 58th Annual Meeting of the Association
  for Computational Linguistics}, pages 7811--7818, Online. Association for
  Computational Linguistics.

\bibitem[{Kuhn and De~Mori(1990)}]{kuhn1990cache}
Roland Kuhn and Renato De~Mori. 1990.
\newblock A cache-based natural language model for speech recognition.
\newblock \emph{IEEE transactions on pattern analysis and machine
  intelligence}, 12(6):570--583.

\bibitem[{Levy(2008)}]{levy2008}
Roger Levy. 2008.
\newblock \href
  {https://doi.org/http://dx.doi.org/10.1016/j.cognition.2007.05.006}
  {Expectation-based syntactic comprehension}.
\newblock \emph{Cognition}, 106(3):1126 -- 1177.

\bibitem[{Levy(2018)}]{levy2018}
Roger~P Levy. 2018.
\newblock \href {https://doi.org/10.31234/osf.io/4cgxh} {Communicative
  efficiency, uniform information density, and the rational speech act theory}.

\bibitem[{Liu(2008)}]{Liu2008}
Haitao Liu. 2008.
\newblock \href {http://www.lingviko.net/JCS.pdf} {Dependency distance as a
  metric of language comprehension difficulty}.
\newblock \emph{Journal of Cognitive Science}, 9(2):159--191.

\bibitem[{Liu et~al.(2017)Liu, Xu, and Liang}]{Liu2017}
Haitao Liu, Chunshan Xu, and Junying Liang. 2017.
\newblock \href {https://doi.org/https://doi.org/10.1016/j.plrev.2017.03.002}
  {Dependency distance: A new perspective on syntactic patterns in natural
  languages}.
\newblock \emph{Physics of Life Reviews}, 21:171 -- 193.

\bibitem[{Luka and Barsalou(2005)}]{luka2005structural}
Barbara~J Luka and Lawrence~W Barsalou. 2005.
\newblock Structural facilitation: Mere exposure effects for grammatical
  acceptability as evidence for syntactic priming in comprehension.
\newblock \emph{Journal of Memory and Language}, 52(3):436--459.

\bibitem[{Meister et~al.(2021)Meister, Pimentel, Haller, J{\"a}ger, Cotterell,
  and Levy}]{meister-etal-2021-revisiting}
Clara Meister, Tiago Pimentel, Patrick Haller, Lena J{\"a}ger, Ryan Cotterell,
  and Roger Levy. 2021.
\newblock \href {https://doi.org/10.18653/v1/2021.emnlp-main.74} {Revisiting
  the {U}niform {I}nformation {D}ensity hypothesis}.
\newblock In \emph{Proceedings of the 2021 Conference on Empirical Methods in
  Natural Language Processing}, pages 963--980, Online and Punta Cana,
  Dominican Republic. Association for Computational Linguistics.

\bibitem[{Misra et~al.(2020)Misra, Ettinger, and Rayz}]{misra2020exploring}
Kanishka Misra, Allyson Ettinger, and Julia Rayz. 2020.
\newblock Exploring bert’s sensitivity to lexical cues using tests from
  semantic priming.
\newblock In \emph{Findings of the Association for Computational Linguistics:
  EMNLP 2020}, pages 4625--4635.

\bibitem[{Petrov et~al.(2006)Petrov, Barrett, Thibaux, and Klein}]{bkp2006}
Slav Petrov, Leon Barrett, Romain Thibaux, and Dan Klein. 2006.
\newblock \href {https://doi.org/10.3115/1220175.1220230} {Learning accurate,
  compact, and interpretable tree annotation}.
\newblock In \emph{Proceedings of the 21st International Conference on
  Computational Linguistics and the 44th Annual Meeting of the Association for
  Computational Linguistics}, ACL-44, pages 433--440, Stroudsburg, PA, USA.
  Association for Computational Linguistics.

\bibitem[{Pickering and Branigan(1998)}]{Pickering1998}
Martin~J. Pickering and Holly~P. Branigan. 1998.
\newblock \href {https://doi.org/10.1006/jmla.1998.2592} {{The Representation
  of Verbs: Evidence from Syntactic Priming in Language Production}}.
\newblock \emph{Journal of Memory and Language}, 39(4):633--651.

\bibitem[{Qian and Jaeger(2012)}]{TingJaeger2012}
Ting Qian and T.~Florian Jaeger. 2012.
\newblock \href {https://doi.org/10.1111/j.1551-6709.2012.01256.x} {Cue
  effectiveness in communicatively efficient discourse production}.
\newblock \emph{Cognitive Science}, 36(7):1312--1336.

\bibitem[{Rajkumar et~al.(2016)Rajkumar, van Schijndel, White, and
  Schuler}]{cog:raja}
Rajakrishnan Rajkumar, Marten van Schijndel, Michael White, and William
  Schuler. 2016.
\newblock \href
  {http://www.sciencedirect.com/science/article/pii/S001002771630155X/}
  {Investigating locality effects and surprisal in written english syntactic
  choice phenomena}.
\newblock \emph{Cognition}, 155:204--232.

\bibitem[{Rajkumar and White(2014)}]{llc:raja:mwhite:2014}
Rajakrishnan Rajkumar and Michael White. 2014.
\newblock \href {https://doi.org/10.1111/lnc3.12090} {Better surface
  realization through psycholinguistics}.
\newblock \emph{Language and Linguistics Compass}, 8(10):428--448.
\newblock ISSN: 1749-818X.

\bibitem[{Ranjan et~al.(2019)Ranjan, Agarwal, and
  Rajkumar}]{ranjan-etal-2019-surprisal}
Sidharth Ranjan, Sumeet Agarwal, and Rajakrishnan Rajkumar. 2019.
\newblock \href {https://www.aclweb.org/anthology/W19-2904}
  {\href{https://www.aclweb.org/anthology/W19-2904/}{Surprisal and Interference
  Effects of Case Markers in {H}indi Word Order}}.
\newblock In \emph{Proceedings of the Workshop on Cognitive Modeling and
  Computational Linguistics}, pages 30--42, Minneapolis, Minnesota. Association
  for Computational Linguistics.

\bibitem[{Ranjan et~al.(2022{\natexlab{a}})Ranjan, Rajkumar, and
  Agarwal}]{cog:sid}
Sidharth Ranjan, Rajakrishnan Rajkumar, and Sumeet Agarwal. 2022{\natexlab{a}}.
\newblock \href
  {https://doi.org/https://doi.org/10.1016/j.cognition.2021.104959} {Locality
  and expectation effects in hindi preverbal constituent ordering}.
\newblock \emph{Cognition}, 223:104959.

\bibitem[{Ranjan et~al.(2022{\natexlab{b}})Ranjan, van Schijndel, Agarwal, and
  Rajkumar}]{ranjan2022dmph}
Sidharth Ranjan, Marten van Schijndel, Sumeet Agarwal, and Rajakrishnan
  Rajkumar. 2022{\natexlab{b}}.
\newblock Dual mechanism priming effects in hindi word order.
\newblock In \emph{Proceedings of the 2nd Conference of the Asia-Pacific
  Chapter of the Association for Computational Linguistics and the 12th
  International Joint Conference on Natural Language Processing,
  {AACL/IJCNLP}}.

\bibitem[{Reitter et~al.(2011)Reitter, Keller, and Moore}]{reitter2011}
David Reitter, Frank Keller, and Johanna~D. Moore. 2011.
\newblock \href {https://doi.org/10.1111/j.1551-6709.2010.01165.x} {A
  computational cognitive model of syntactic priming}.
\newblock \emph{Cognitive Science}, 35(4):587--637.

\bibitem[{Sangal et~al.(1995)Sangal, Chaitanya, and
  Bharati}]{sangal1995natural}
Rajeev Sangal, Vineet Chaitanya, and Akshar Bharati. 1995.
\newblock \emph{Natural language processing: a Paninian perspective}.
\newblock PHI Learning Pvt. Ltd.

\bibitem[{Scheepers(2003)}]{Scheepers2003}
C~Scheepers. 2003.
\newblock Syntactic priming of relative clause attachments: persistence of
  structural configuration in sentence production.
\newblock \emph{Cognition}, 89:179--205.

\bibitem[{Staub(2015)}]{staub2015}
Adrian Staub. 2015.
\newblock \href {https://doi.org/10.1111/lnc3.12151} {The effect of lexical
  predictability on eye movements in reading: Critical review and theoretical
  interpretation}.
\newblock \emph{Language and Linguistics Compass}, 9(8):311--327.

\bibitem[{Stolcke(2002)}]{SRILM-ICSLP:2002}
Andreas Stolcke. 2002.
\newblock {SRILM} --- {A}n extensible language modeling toolkit.
\newblock In \emph{Proc.\ ICSLP-02}.

\bibitem[{Temperley(2007)}]{Temperley2007}
David Temperley. 2007.
\newblock \href {https://doi.org/DOI: 10.1016/j.cognition.2006.09.011}
  {Minimization of dependency length in written {E}nglish}.
\newblock \emph{Cognition}, 105(2):300--333.

\bibitem[{Tooley and Traxler(2010)}]{tooley2010syntactic}
Kristen~M Tooley and Matthew~J Traxler. 2010.
\newblock Syntactic priming effects in comprehension: A critical review.
\newblock \emph{Language and Linguistics Compass}, 4(10):925--937.

\bibitem[{van Schijndel and Linzen(2018)}]{van2018neural}
Marten van Schijndel and Tal Linzen. 2018.
\newblock A neural model of adaptation in reading.
\newblock In \emph{Proceedings of the 2018 Conference on Empirical Methods in
  Natural Language Processing}, pages 4704--4710.

\bibitem[{Vasishth(2004)}]{vasishthysall04}
S.~Vasishth. 2004.
\newblock \href
  {http://www.ling.uni-potsdam.de/~vasishth/pdfs/Vasishth2004YearbookSAL.pdf}
  {Discourse context and word order preferences in {H}indi}.
\newblock \emph{Yearbook of South Asian Languages}, pages 113--127.

\bibitem[{Wells et~al.(2009)Wells, Christiansen, Race, Acheson, and
  MacDonald}]{wells2009experience}
Justine~B Wells, Morten~H Christiansen, David~S Race, Daniel~J Acheson, and
  Maryellen~C MacDonald. 2009.
\newblock Experience and sentence processing: Statistical learning and relative
  clause comprehension.
\newblock \emph{Cognitive psychology}, 58(2):250--271.

\bibitem[{Wierzba and Fanselow(2020)}]{wierzba2020factors}
Marta Wierzba and Gisbert Fanselow. 2020.
\newblock Factors influencing the acceptability of object fronting in german.
\newblock \emph{The Journal of Comparative Germanic Linguistics},
  23(1):77--124.

\bibitem[{Yadav et~al.(2017)Yadav, Vaidya, and Husain}]{Yadav2017KeepingIS}
Himanshu Yadav, Ashwini Vaidya, and Samar Husain. 2017.
\newblock Keeping it simple: Generating phrase structure trees from a {H}indi
  dependency treebank.
\newblock In \emph{TLT}.

\end{thebibliography}

\pagebreak

\appendix

\section*{Appendix}
\section{Variant Generation}\label{appendix:A}

\begin{figure*}[t]
\centering
\noindent\makebox[\textwidth]{%
\subfloat[Figure][Dependency tree]{
\label{fig:intro-tree}
\includegraphics[scale=0.55]{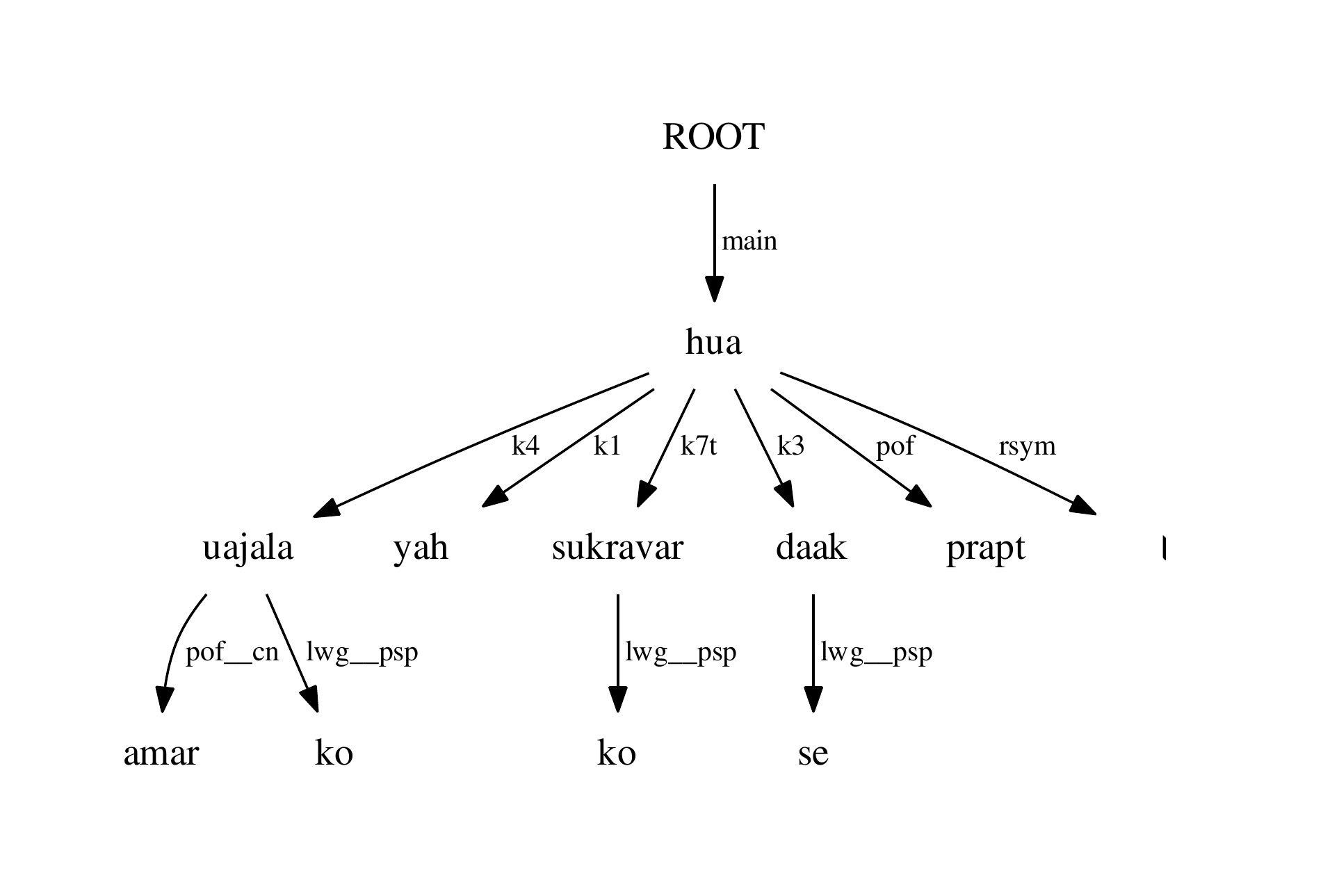}
}
\subfloat[Table][Dependency relations]{
\label{tab:intro-tree-labels}
  \begin{scriptsize}
\begin{tabular}[b]{lc}
\hline 
Label & Dependency \\
      & relation\\\hline
{\it Invariant syntactic relations}&\\
k1 & subject/agent\\
k2 & object/patient\\
k3 & instrument\\
k4 & object/recipient\\
k7t & location in time\\\hline
{\it Complex predicate relation}&\\
pof & parts of\\
          & conjunct verb\\
pof\_cn &  parts of\\
          & compound noun\\\hline
{\it Local word group (lwg)} &\\
lwg\_psp & postposition\\
lwg\_vaux & auxilliary verb\\\hline
{\it Symbols} &\\
rsym & symbol relation\\
\end{tabular}
   \end{scriptsize}
}}
\caption{Example HUTB dependency tree and relation labels}
\label{fig:hutb-tree}
\end{figure*}

\begin{exe}

  \ex 
  {\label{ex:hindi-prev-sent-app} \textbf{Context sentence}
    \gll {\underline{amar ujala}-ki} {bhumika} {nispaksh} {rehti} {hai}\\
         {Amar Ujala-\textsc{gen}} {role} {unbiased} {remain} {be\textsc{.prs.sg}}\\
     \glt Amar Ujala's role remains unbiased.}  

\end{exe}

\begin{exe}

  \ex \label{ex:hindi-intro-app}
  \begin{xlist}
  \ex[]{\label{ex:hindi-order-ref-app}
    \gll {\underline{amar ujala}-ko} {\bf yah} {\emph{sukravar}-ko} {daak-se} {prapt} {hua} [Given-Given = 0] \textbf{(Reference)}\\
         {\underline{Amar Ujala}-\textsc{acc}} {\bf it} {\emph{friday}-on} {post-\textsc{inst}} {receive} {be\textsc{.pst.sg}}\\
     \glt \underline{Amar Ujala} received \textbf{it} by post on \emph{Friday}.}
  \ex[] {\label{ex:hindi-order-var1-app} {{\bf yah} \underline{{amar ujala}}-ko \emph{sukravar}-ko daak-se prapt hua} [Given-Given = 0] \textbf{(Variant 1)}}
  \ex[] {\label{ex:hindi-order-var2-app} {\emph{sukravar}-ko {\bf yah} \underline{amar ujala}-ko daak-se prapt hua} [New-Given = -1] \textbf{(Variant 2)}}
  \end{xlist}

\end{exe}

This work uses sentences from the Hindi-Urdu Treebank (HUTB) corpus of dependency trees~\citep{Bhatt2009} containing well-defined subject and object constituents. Figure \ref{fig:hutb-tree} displays the dependency tree (and a glossary of relation labels) for reference sentence \ref{ex:hindi-order-ref-app}. The grammatical variants were created using an algorithm that took as input the dependency tree corresponding to each HUTB reference sentence. The re-ordering algorithm permuted the preverbal\footnote{Hindi is not a strictly verb-final language but the majority of the constituents in the HUTB corpus are preverbal. Our corpus analysis of 13274 sentences present in HUTB suggests 20,750 pairs of preverbal constituents and 2599 pairs of postverbal constituents. Therefore, our variant generation (via reordering of constituents) and subsequent experiments focus on word-order variation in the preverbal domain, considering preverbal domain to be the locus of word order variation. Only preverbal constituents are permuted to generate grammatical variants and leave the postverbal constituents in the reference-variants sentences as it is.} dependents of the root verb and linearized the resulting tree to obtain variant sentences. For example, corresponding to the reference sentence \ref{ex:hindi-order-ref-app} and its root verb ``hai'' (see figure \ref{fig:intro-tree}), the preverbal constituents with parents as ``ujala'', ``yah'', ``suravar'', ``daak'', and ``prapt'' were permuted to generate the artificial variants (\ref{ex:hindi-order-var1-app} and \ref{ex:hindi-order-var2-app}). The ungrammatical variants were automatically filtered out using dependency relation sequences (denoting grammar rules) attested in the gold standard corpus of HUTB trees. In the dependency tree \ref{fig:intro-tree}, ``k4-k1'', ``k7t-k1'', ``k3-k7t'', and ``pof-k3'' are dependency relation sequences. In cases where the total number of variants exceeded 100 (a random cutoff),\footnote{Higher and lower cutoffs do not affect our results.} we chose 99 non-reference variants randomly along with the reference sentence.\\

\section{Information Status Annotation}\label{appendix:B}

The subject and object constituents in a sentence were assigned a \textit{Given} tag if any content word within them was mentioned in the preceding sentence or if the head of the phrase was a pronoun. All other phrases were tagged as \textit{New}. The sentence example \ref{ex:hindi-intro-app} illustrates the proposed annotation scheme. 

\begin{itemize}

\item Example \ref{ex:hindi-order-ref-app} follows {\it Given-Given} ordering --- The object ``Amar Ujala'' in the sentence is mentioned in the preceding context sentence \ref{ex:hindi-prev-sent-app}, it would be annotated as \emph{Given}. In contrast, the subject ``yah'' is a pronoun so it would also be tagged as \emph{Given} following the annotation scheme. 

\item Example \ref{ex:hindi-order-var2-app} follows {\it New-Given} ordering --- The object ``sukravar" in the sentence should be tagged as \emph{New} as it is not mentioned in the preceding context sentence \ref{ex:hindi-prev-sent-app}. In contrast, the subsequent pronoun ``yah'' which acts as the subject of the sentence should be tagged as \emph{Given} following the annotation scheme. 

\end{itemize}

\section{Correlation Plot}\label{appendix:C}
\begin{figure*}[!htbp]
    \centering
    \scalebox{0.7}{
    \includegraphics{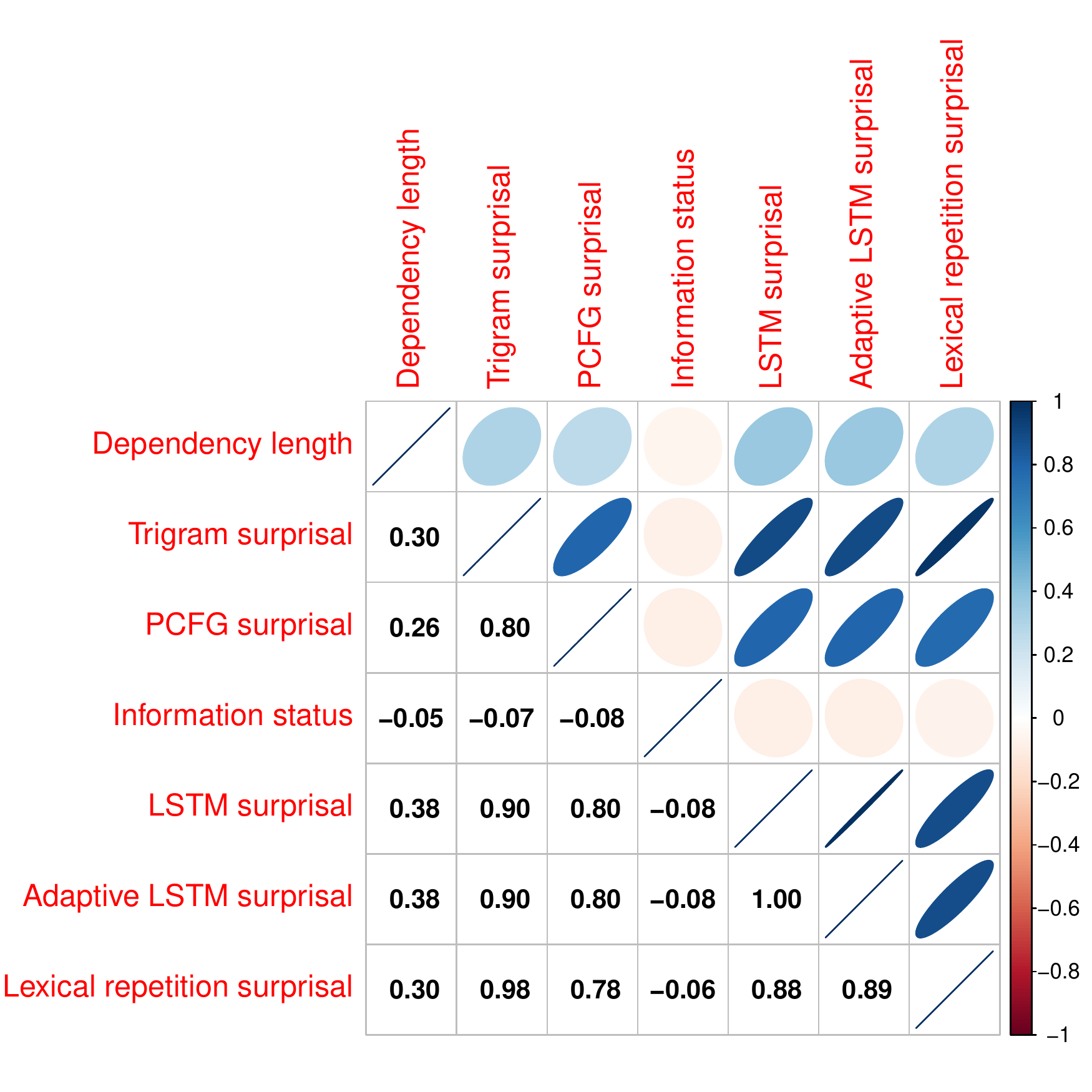}}
    \caption{Pearson’s coefficient
of correlation between different pairs of predictors}
    \label{fig:corr-plot}
\end{figure*}

The Pearson's correlation coefficients between different predictors are displayed in Figure \ref{fig:corr-plot}. The adaptive LSTM surprisal has a high correlation with all other surprisal features and a low correlation with dependency length and information status score.

\section{Joachims Transformation} \label{appendix:joc}

This technique converts a binary classification problem into a pair-wise ranking task involving the feature vectors of a reference sentence and each of its variants. Table \ref{tab:joc-trans} displays the \citeauthor{Joachims:2002}'s transformation. The delta ($\delta$) refers to the difference between feature vectors of the reference sentence and its paired variant. The overall goal is to model two-alternative choices for each reference sentence such that the speaker generated the reference sentence after rejecting a potential grammatical variant. Moreover, the reference sentence that appeared originally in the corpus must have been present due to its properties (\emph{viz.,} dependency length, discourse context, accessibility, or surprisal), and variant sentences are less likely to be produced due to the same reasons.

\begin{table*}[ht]
\centering
\subfloat[Subtable 1 list of tables text][Original feature values]{
\scalebox{0.8}{
\begin{tabular}{|c|c|ccc|}
\hline
Condition& Label & Dependency & $n$-gram  & pcfg\\
         &       & length     & surprisal & surprisal\\\hline
Reference   &  1 & 18         & 24.69  &  61.13\\\hline
Variant$_1$ &  0 & 20         & 23.80  &  60.67\\\hline
Variant$_2$ &  0 & 18         & 23.02  &  60.02\\\hline
\end{tabular}}  
\label{tab:orig}
}
\subfloat[Subtable 2 list of tables text][Transformed feature values]{
\scalebox{0.8}{
\begin{tabular}{|c|c|ccc|}
\hline
Condition& Label & $\delta$ Dependency & $\delta$ $n$-gram & $\delta$ pcfg\\
         &       & length              & surprisal         & surprisal\\\hline
Variant$_1$      &  0  &  2  & -0.89  &  -0.46\\
-Reference       &     &     &       &  \\\hline
Reference        &  1  &  0 & 1.67 &  1.11\\
-Variant$_2$     &     &     &       &  \\\hline
\end{tabular}}
\label{tab:trans}  
}
\caption{Joachims transformation}
  \label{tab:joc-trans}  
\end{table*}

\section{Information Profile for IO-fronted Example}\label{appendix:io-fronting}

The LSTM LM when adapted to the previous sentence (\ref{ex:hindi-prev-sent-app}) in the discourse assigns lower surprisal score to the \textit{given} item when it occurs in the first position (``amar ujala" in sentence \ref{ex:hindi-order-ref-app}) than when it appears in the second position (``amar ujala" in sentence \ref{ex:hindi-order-var1-app}) in the subsequent sentence. 

\begin{figure*}[ht]

    \begin{center}
    \includegraphics[width=1\textwidth]{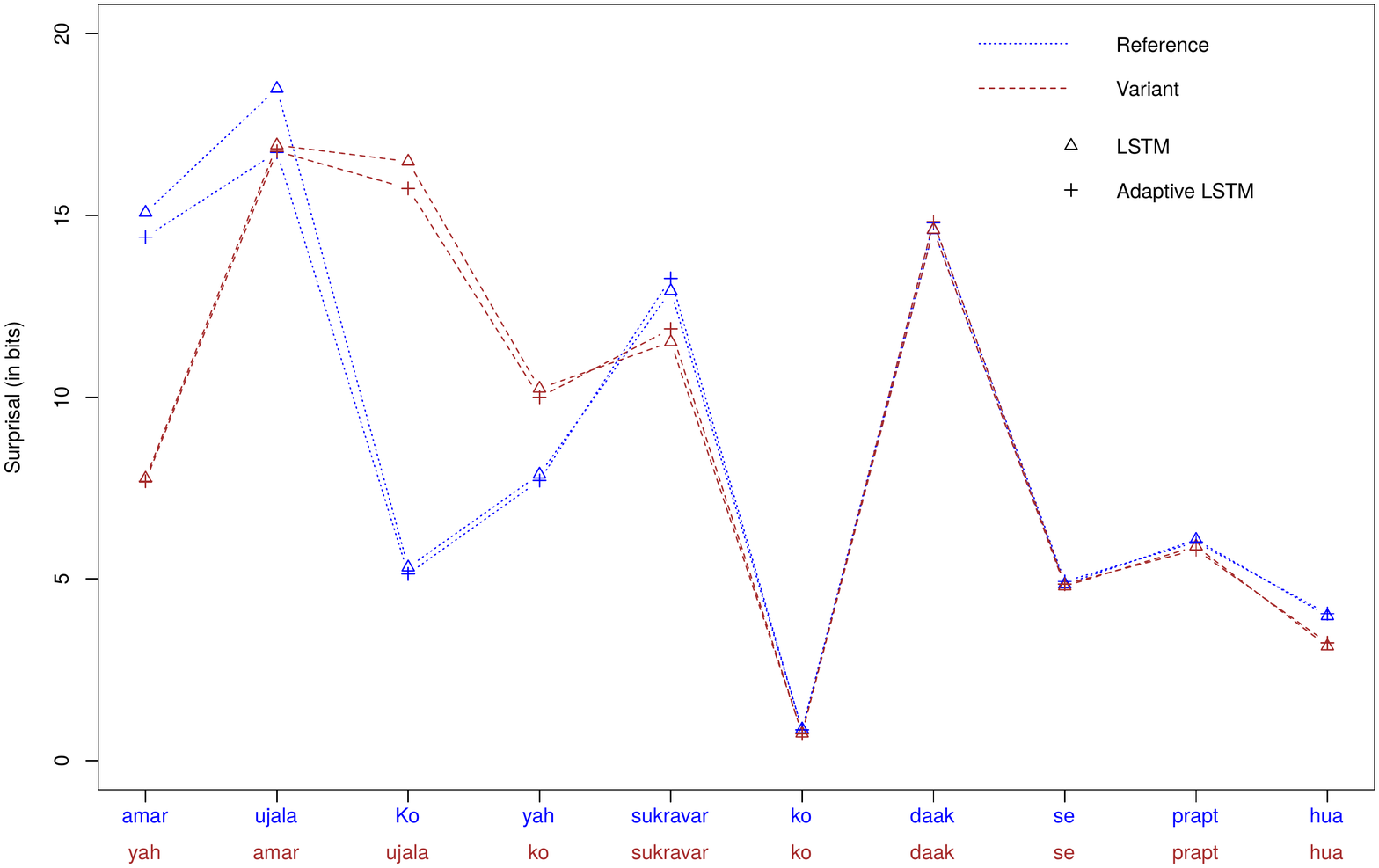}
    \end{center}
    \caption{Information profile for the reference-variant pair \ref{ex:hindi-order-ref-app} and \ref{ex:hindi-order-var1-app}}
    \label{fig:io-profile}
\end{figure*}

\begin{table*}[ht]
\scalebox{0.8}{
\begin{tabular}{lcccccccc}
\multicolumn{2}{c}{Type} & {3g surp} & Deplen & PCFG surp & IS score & LSTM surp & Adaptive LSTM surp & Lex rept surp \\\hline
Example \ref{ex:hindi-order-ref-app} & Reference & 24.69 & 18 & 61.13 & 0 & \textbf{91.80} & \textbf{89.52} & 23.80 \\
Example \ref{ex:hindi-order-var1-app} & Variant & 23.80 & 20 & 60.67 & 0 & 93.78 & 93.17 & 22.19 \\\hline
\end{tabular}}
\caption{Predictor scores for reference-variant pairs (\ref{ex:hindi-order-ref-app}, \ref{ex:hindi-order-var1-app})}
\label{tab:sent-level-scores}
\end{table*}

\section{Contextual Adaptation on One Vs. Multiple Sentences for DO/IO Constructions}\label{appendix:prev5}

\begin{table*}[ht]
\centering
\scalebox{0.8}{
\begin{tabular}{|c|c|c|c|c|c|c|}
\hline
\textbf{\begin{tabular}[c]{@{}c@{}}Non-canonical \\ HUTB Sentences\end{tabular}} & \textbf{\begin{tabular}[c]{@{}c@{}}Frequency \\ Count (\%)\end{tabular}} & \textbf{\begin{tabular}[c]{@{}c@{}}Baseline \\ Perplexity\end{tabular}} & \textbf{\begin{tabular}[c]{@{}c@{}}Adapted \\ Perplexity (Prev1)\end{tabular}} & \textbf{\begin{tabular}[c]{@{}c@{}}Perplexity \\ Dip (Prev1)\end{tabular}} & \textbf{\begin{tabular}[c]{@{}c@{}}Adapted \\ Perplexity (Prev5)\end{tabular}} & \textbf{\begin{tabular}[c]{@{}c@{}}Perplexity \\ Dip (Prev5)\end{tabular}} \\ \hline
\textbf{DO} & 133 (1\%) & 183.92 & 103.40 & -80.52 & 77.33 & -106.59 \\ \hline
\textbf{IO} & 101 (0.76\%) & 138.78 & 88.26 & -50.52 & 68.45 & -70.33\\ \hline
\end{tabular}}
\caption{Effect of adaptation on discourse sentences (Prev1: Preceding one sentence in discourse, Prev5: Preceding five sentences in discourse)}
\label{tab:lex-adapt-lexical}
\end{table*}    

We investigated if adapting the LSTM LM to the preceding five contextual sentences instead of one contextual sentence will help predict word-ordering patterns better for IO/DO constructions. Table \ref{tab:lex-adapt-lexical} showcases perplexity dip on test sentences during 1 vs. 5 contextual sentence adaptation. Table \ref{tab:3vs5:pred} highlights classification accuracy of different models containing combination of features. Our results indicate that adding \textit{Prev5-adaptive} LSTM surprisal in the machine learning model above and beyond every other features including \textit{Prev1-adaptive} surprisal (\emph{i.e.,} Baseline) does not significantly boost prediction accuracy for both IO- and DO-fronted subset. A similar finding was observed when Prev5-lexical-repetition surprisal (\textit{Prev5-Lex-Rept LM}: base trigram LM interpolated with unigram cache LM containing 5 preceding sentences history) was included in the classifier model above and beyond every other features (\emph{i.e.,} Baseline) including \textit{Prev1-lexical-repetition} surprisal (\textit{Prev1-Lex-Rept LM}: base trigram LM interpolated with unigram cache LM containing only 1 preceding sentence history).

\begin{table*}[ht]
\centering
\begin{tabular}{l|ccc}
  \textbf{Type} & \textbf{Baseline} & \textbf{+ Prev5 Adaptive} & \textbf{+ Prev5 Lex-Rept} \\
                &   & \textbf{  LSTM Surp} & \textbf{  Surp} \\
                      \toprule
{DO} & 81.06 & 81.12 & 80.94\\
{IO} & 89.65 & 89.73 & 89.51\\\bottomrule

\end{tabular}
\caption{Prediction performance (Direct objects (DO: 1663 points), Indirect Objects (IO: 1353 points)); Baseline denotes \textit{base1+g} shown in Table \ref{tab:lex-adapt-pred-acc}; bold denotes McNemar's two-tailed significance compared to baseline model in the same row}
\label{tab:3vs5:pred}
\end{table*}

\section{Human Evaluation}\label{subsec:human-eval}

To determine whether the permuted word order (variant) is dispreferred to the original word order (reference), we conducted a targeted human evaluation via forced-choice rating task and collected sentence ratings from 12 Hindi native speakers for 167 randomly selected reference-variant pairs in our data set. Participants were first shown the preceding sentence, and then they were asked to judge the subsequent most likely sentence as the best choice between the reference-variant pair. 
Each sentence was assigned a human label of ``1'' if more than 50\% participants voted it as best choice else human label of ``0''. The stimuli belonged to two different constructions, viz., the reference sentence (Ref) has canonical ordering whereas, the variant (Var) has non-canonical ordering (DO-fronted or IO-fronted) and vice versa.

Table \ref{tab:human-eval} presents the results. On the entire dataset containing 167 reference-variant pairs, 89.92\% (agreement accuracy) of the reference sentences which originally appeared in the HUTB corpus were also preferred by native speakers compared to the artificially generated grammatical variants expressing similar meaning. Overall, the full model containing all the features, including adaptive LSTM surprisal predicted human preference (76.65\%) much better than corpus choice labels (74.85\%). Furthermore, the Pearson's correlation coefficient of the classifier's prediction with the human judgement is 0.534 and with corpus labels is 0.497. Interestingly, as initially hypothesized, the Hindi participants were more prone to prefer IO-fronted construction (80\%) compared to DO-fronted construction (65\%) as captured by the agreement accuracy validating the findings reported in Table \ref{tab:lex-adapt-pred-acc}.

\begin{table*}[ht]
    \centering
    \scalebox{1}{
    \begin{tabular}{l|ccc}
    Construction & Agreement (\%) & Model (\%) & Model (\%) \\
    Type (item count) & human:corpus & corpus labels & human labels\\\hline
    Ref: Canonical | Var: DO-fronted (20) & 95 & 90 & 85\\
    \textbf{Ref: DO-fronted | Var: Canonical} (20) & \textbf{65} & 25 & 50\\\hline
    Ref: Canonical | Var: IO-fronted (20) & 100 & 85 & 85\\
    \textbf{Ref: IO-fronted | Var: Canonical} (20) & \textbf{80} & 65 & 65\\\hline
    Ref: Non-canonical | Var: Canonical   (80) & 85.00 & 66.25 & 71.25\\
    Ref: Canonical | Var: Canonical   (87) & 94.25 & 82.76 & 81.61\\\hline
    Total (167) & 89.92 & 74.85 & 76.65\\\hline
    \end{tabular}}
    \caption{Targeted human evaluation --- \textbf{Agreement human/corpus}: Percentages of times human judgement matches with corpus reference choice; \textbf{Model corpus}: Percentages of corpus choice correctly predicted by the classifier containing all the predictors (\emph{base1 + g} as per Table \ref{tab:lex-adapt-pred-acc}); \textbf{Model human:} Percentages of human label correctly predicted by the classifier containing all the predictors (\emph{base1 + g} as per Table \ref{tab:lex-adapt-pred-acc})}
    \label{tab:human-eval}
\end{table*}

\section{Variance Inflation Factor Analysis}\label{subsec:vif}

Table \ref{tab:orig-vif} displays the VIF scores for each predictors in the different regression models. The VIF scores for the regression models without containing the correlated features, such as trigram surprisal and vanilla LSTM surprsial is documented in Table \ref{tab:trans-vif}. Correspondingly, Table \ref{tab:regr-results-vif} reports the results of regression experiment when the model did not containing these highly correlated features.

\begin{table*}[ht]
\subfloat[Subtable 1 list of tables text][All predictors]{
\scalebox{0.71}{
\begin{tabular}{l|ccc}
\hline
\textbf{Predictors} & \textbf{Full (72833)} & \textbf{DO (1663)} & \textbf{IO (1353)} \\ \hline
trigram surp & \textbf{27.61} & 18.87 & 18.87 \\
dependency length & 1.18 & 1.46 & 1.46 \\
pcfg surp & 3.08 & 1.86 & 1.86 \\
IS score & 1.01 & 1.03 & 1.03 \\
lex-rept surp & \textbf{24} & 16.47 & 16.47 \\
lstm surp & \textbf{241.62} & 109.04 & 109.04 \\
adaptive lstm surp & \textbf{244.37} & 106.97 & 106.97 \\\hline
\textsc{performance}& & &\\
Residual standard err & 0.271 & 0.294 & 0.354 \\
Multiple R-squared & 0.707 & 0.657 & 0.502 \\
Adjusted R-squared & 0.707 & 0.655 & 0.500 \\\hline
\end{tabular}
}  
\label{tab:orig-vif}
}
\subfloat[Subtable 2 list of tables text][Predictors except trigram and lstm surprisal measures]{
\scalebox{0.82}{
\begin{tabular}{l|ccc}
\hline
\textbf{Predictors} & \textbf{Full (72833)} & \textbf{DO (1663)} & \textbf{IO (1353)} \\ \hline
dependency len & 1.18 & 1.42 & 1.18 \\
pcfg surp & 2.95 & 1.73 & 2.08 \\
IS score & 1.01 & 1.01 & 1.06 \\
lex-rept surp & 4.99 & 4.45 & 4.26 \\
adaptive lstm surp & 5.77 & 4.39 & 4.29 \\\hline
\textsc{performance}& & &\\
Residual std. err & 0.271 & 0.357 & 0.304 \\
Multiple R-sqrd & 0.706 & 0.492 & 0.633 \\
Adjusted R-sqrd & 0.706 & 0.490 & 0.631 \\ \hline
\end{tabular}
}
\label{tab:trans-vif}  
}

\caption{Variance inflation factor analysis on different regression models containing: (a) all predictors b) all predictors but correlated features; Each column denotes individual models on a given dataset with different set of predictors; VIF larger than 5 or 10 indicates that the model has problems estimating the coefficient of variables} 
\label{tab:vif-score}
\end{table*}

\begin{table*}
\centering
\scalebox{1}{
\begin{tabular}{lccc}
\toprule
Predictor & $\hat\beta$ & $\hat\sigma$ & t\tabularnewline
\midrule 
Intercept & 1.5 & 0.001 & 1493.53 \\
dependency len & 0.02 & 0.0011 & 15.84 \\
pcfg surp & -0.08 & 0.0017 & -43.49 \\
IS score & 0.01 & 0.001 & 11.92 \\
lex-rept surp & -0.09 & 0.0022 & -39.97 \\
adaptive lstm surp & -0.28 & 0.0024 & -116.06 \\
\bottomrule 
\end{tabular}}
\caption{Regression model on full data set after removing the correlated features ($N=72833$; all significant predictors denoted by $|$t$|$\textgreater{}2)}
\label{tab:regr-results-vif}
\end{table*}

\end{document}